\documentclass[letterpaper]{article} 
\usepackage{aaai23}  
\nocopyright
\usepackage{times}  
\usepackage{helvet}  
\usepackage{courier}  
\usepackage[hyphens]{url}  
\usepackage{graphicx} 
\usepackage{natbib}  
\usepackage{caption} 
\usepackage{algorithm}
\usepackage{algorithmic}

\usepackage{newfloat}
\usepackage{listings}
\usepackage{amssymb}

\usepackage{booktabs} 
\usepackage[T1]{fontenc}
\usepackage{multirow}
\usepackage{subcaption}

\DeclareCaptionStyle{ruled}{labelfont=normalfont,labelsep=colon,strut=off} 
\floatstyle{ruled}
\newfloat{listing}{tb}{lst}{}
\floatname{listing}{Listing}
\title{Mask-based Predictive Representations for Reinforcement Learning}
\author{
    Kai Zhao\textsuperscript{\rm 1}\thanks{Work completed at Beijing Normal University in late 2023.}
}
\affiliations{
    \textsuperscript{\rm 1}School of Systems Science, Beijing Normal University\\
    Beijing, China
}

\usepackage{bibentry}

\begin{document}

\maketitle

\begin{abstract}
Vision-based deep reinforcement learning involves dealing with high-dimensional inputs of image information.   It is crucial to abstract effective states from high-dimensional image inputs and limited samples for sample-efficient reinforcement learning.  
To address this challenge, inspired by fields such as natural language processing and computer vision, we propose a self-supervised task based on mask prediction as an auxiliary task for reinforcement learning. 
This non-reconstruction method uses the sequence information collected by the agent from the environment and the context information in the sequence to predict the masked information, thereby strengthening the agent's understanding of the task and learning effective representations.
Combined with transformers, we find that the model reconstructs the masked input sequence in the latent space.   By feeding the compressed representations learned by this method into reinforcement learning models, we observe an improvement in the sample efficiency of reinforcement learning.   Moreover, the model outperforms state-of-the-art sample-efficient reinforcement learning methods on multiple continuous and discrete control benchmarks.

\end{abstract}

\section{Introduction}
Deep reinforcement learning (DRL) has made significant progress in sequential decision-making problems, often surpassing human performance in many scenarios \cite{tassa2018deepmind, berner2019dota, bellemare2013arcade}. 
However, the learning process of reinforcement learning (RL) requires a substantial amount of data resources \cite{schrittwieser2020mastering, badia2020agent57}. 
Collecting a large amount of training data in the real world is costly and often poses risks \cite{dulac2021challenges}. Therefore, data-efficient reinforcement learning is crucial.
Utilizing pixel data to learn effective state representations has proven to be a potent strategy for achieving data-efficient reinforcement learning, as demonstrated by \citet{schwarzer2020data}.
Inspired by the success of mask-based pretraining in the fields of natural language processing (NLP) \cite{devlin2018bert, brown2020language} and computer vision(CV) \cite{he2022masked, assran2023self}, we make the endeavor to explore the idea of mask-based modeling in RL.


In natural language processing and computer vision, reconstructing the original input through masking modeling is an effective pre-training method \cite{he2022masked, devlin2018bert}.
In reinforcement learning, a critical unresolved issue is how to learn effective state representations for agents. While pre-training methods can learn state representations in advance, in reinforcement learning, agents continuously update their own experiences through interaction with the environment. Pre-trained models cannot be updated along with the agent as it evolves.
Additionally, most pre-training methods learn compressed representations by reconstructing the pixel space \cite{he2022masked}.  In reinforcement learning, the observation space is often characterized by high dimensionality and the presence of redundant information. Compressing and feature learning in the image space can help agents improve their strategies more quickly and effectively \cite{yarats2021improving, DBLP:conf/iclr/LiuZZQZLYL21, cetin2022stabilizing, tang2023understanding}.

Building upon the analysis presented earlier, and directly inspired by \citet{assran2023self, yu2022mask}, we introduce the Mask-based Predictive Representation (MPR) learning method. MPR is an effective self-supervised learning technique designed for reconstructing latent spaces within vision-based reinforcement learning frameworks. MPR coordinates the objectives of representation learning and policy learning for joint optimization. Importantly, it reconstructs the masked image sequence within the latent space using contextual cues provided by continuous frames, rather than the input space. This strategic choice effectively reduces the computational complexity usually associated with pixel-level reconstruction.


In reinforcement learning, there is a high degree of correlation between consecutive frames. The replay buffer disrupts the correlation between sequences to train a more robust policy \cite{mnih2013playing}. However, MPR leverage thecorrelation between consecutive frames to enhance the informativeness, predictiveness, and consistency of the learned representations in both spatial and temporal dimensions. This continuous frame prediction enhances the agent's awareness of the global context information in the entire input observations and encourages the state representations to be more predictive in spatial and temporal dimensions. The encoding of global predictive information is encouraged into each frame-level state representation, achieving better representation learning and further promoting policy learning.

To summarize, our contributions include:
\begin{itemize}
	%
	\item We introduce mask-based sequence modeling into reinforcement learning, leveraging self-predictive learning to acquire adaptive representations for various tasks, thus enhancing the sample efficiency of RL.

	\item We propose the mask-based predictive representation RL model, which utilizes contextual information among observation sequences to reconstruct masked information in each image within the latent space, thereby facilitating policy learning for agents.

	\item Empirical experiments demonstrate that MPR achieves leading performance across multiple continuous control tasks. By applying MPR to the Atari game benchmark, we achieve top results in 11 out of 26 games, indicating that MPR enables sample-efficient reinforcement learning."
\end{itemize}

\section{Related Work}

\subsection{Representation learning for RL}
Training reinforcement learning agents directly from visual inputs has advantages such as sample efficiency, task relevance, and high practical value. However, learning reinforcement learning policies directly from high-dimensional state-action spaces also faces significant challenges due to the presence of redundant information in these spaces \cite{shelhamer2016loss}. Recent research has addressed this challenge by employing self-supervised learning methods \cite{hessel2018rainbow, laskin2020curl, liu2022data, yu2022mask, tang2023understanding}. A common approach is to learn high-quality image representations through auxiliary tasks while learning the main policy objective. Depending on the auxiliary task, these methods can be categorized into tasks such as pixel reconstruction \cite{yarats2021improving}, bisimulation \cite{zhang2020learning, allen2021learning}, and dynamics prediction \cite{guo2020bootstrap, lee2020stochastic, schwarzer2020data}. Such tasks primarily utilize loss functions in the style of BYOL \cite{DBLP:conf/nips/GrillSATRBDPGAP20} or CURL \cite{laskin2020curl} to learn high-quality state representations without the need for negative samples. Some works also employ pretraining to learn state representations, but they often require sample collection in advance and may not adapt well to specific tasks. Our goal is to design an auxiliary representation learning task that updates concurrently with policy learning to improve the sample efficiency of reinforcement learning.

\subsection{Sample-efficient Reinforcement Learning}

Collecting trajectories in the real world is expensive and dangerous. Using simulated environments for trajectory collection can provide a large amount of training data for training agents, but it comes with the cost of increased computational overhead. Therefore, developing sample-efficient reinforcement learning (RL) training methods is crucial. In RL, sample efficiency refers to the finite number of interactions between the agent and the environment. High sample efficiency means that the agent can improve its policy using a limited number of trajectories. Additionally, representation learning can help address the problem of high-dimensional state spaces in reinforcement learning. To improve the sample efficiency of vision-based reinforcement learning, recent research has introduced self-supervised representation learning into reinforcement learning, improving learned representations by designing auxiliary tasks \cite{anand2019unsupervised, lee2020stochastic, laskin2020reinforcement, schwarzer2020data, yarats2021improving, yarats2021mastering}. Data augmentation techniques are commonly used in auxiliary tasks to help the agent learn effective representations \cite{yarats2020image, laskin2020reinforcement}, and these approaches fall under the umbrella of model-free methods. Furthermore, there is also a branch of work focusing on world models \cite{hafner2019dream, hafner2019learning, hafner2020mastering, hafner2023mastering, ye2021mastering, deng2022dreamerpro, campbell2019model}, namely model-based reinforcement learning, where the world's transition dynamics are learned in visual or latent space, and planning, imagination, or policy learning is performed based on the acquired models. This paper focuses on the avenue of auxiliary tasks.


\subsection{Masked Modeling in RL}

Masking modeling in reinforcement learning has been extensively studied \cite{zheng2022online}. MVP \cite{xiao2022masked} investigates transferring pretrained visual representations to RL tasks. TT \cite{janner2021offline} explores autoregressive next token prediction for model-based RL applications. MaskDP \cite{liu2022masked} randomly masks a portion of trajectories and generalizes prior masking strategies such as inverse dynamics.
In contrast to the aforementioned works on masking trajectories, inspired by masking modeling in NLP and CV \cite{brown2020language, bao2021beit, he2022masked, yu2021playvirtual}, \citet{yu2022mask} explores mask-based modeling for RL to exploit the high correlation in vision data and enhance agents’ awareness of global-scope dynamics when learning state representations. Building upon this, we investigate the impact of different masking schemes on RL agent learning effectiveness. Furthermore, we utilize a joint prediction framework to study the influence of highly correlated data in different perspectives within masked sequences on reinforcement learning performance.

\section{Approach}
\subsection{Background}
Consider an environment formulated as Markov Decision Process (MDP) defined by a tuple $(\mathcal{S}, \mathcal{A}, T, r, d_{0}, \gamma)$, where $\mathcal{S}$ is the state space, $\mathcal{A}$ is the action space, $T(s'|s, a)$ is the transition probability distribution, $r: \mathcal{S} \times \mathcal{A} \rightarrow \mathbb{R}$ is the reward function, $d_{0}$ is the initial state distribution, and $\gamma \in (0, 1]$ is the discount factor. The goal of reinforcement learning is to find an optimal policy $\pi(\mathbf{a}|\mathbf{s})$ that maximizes the cumulative discounted reward $\mathbb{E}_{s_{t},a_{t}}[\sum^{\infty}_{t=0}\gamma^{t}r(s_{t}, a_{t})]$, where $\mathbf{s}_{0} \sim d_{0}(\cdot)$, $\mathbf{a}_{t} \sim \pi(\cdot|s_{t})$, and $\mathbf{s}_{t+1} \sim T(\cdot|\mathbf{s}_{t},\mathbf{a}_{t})$.

\subsection{Mask-based Predictive Representations}
\begin{figure}[t]
	\centering
	\includegraphics[scale=0.75]{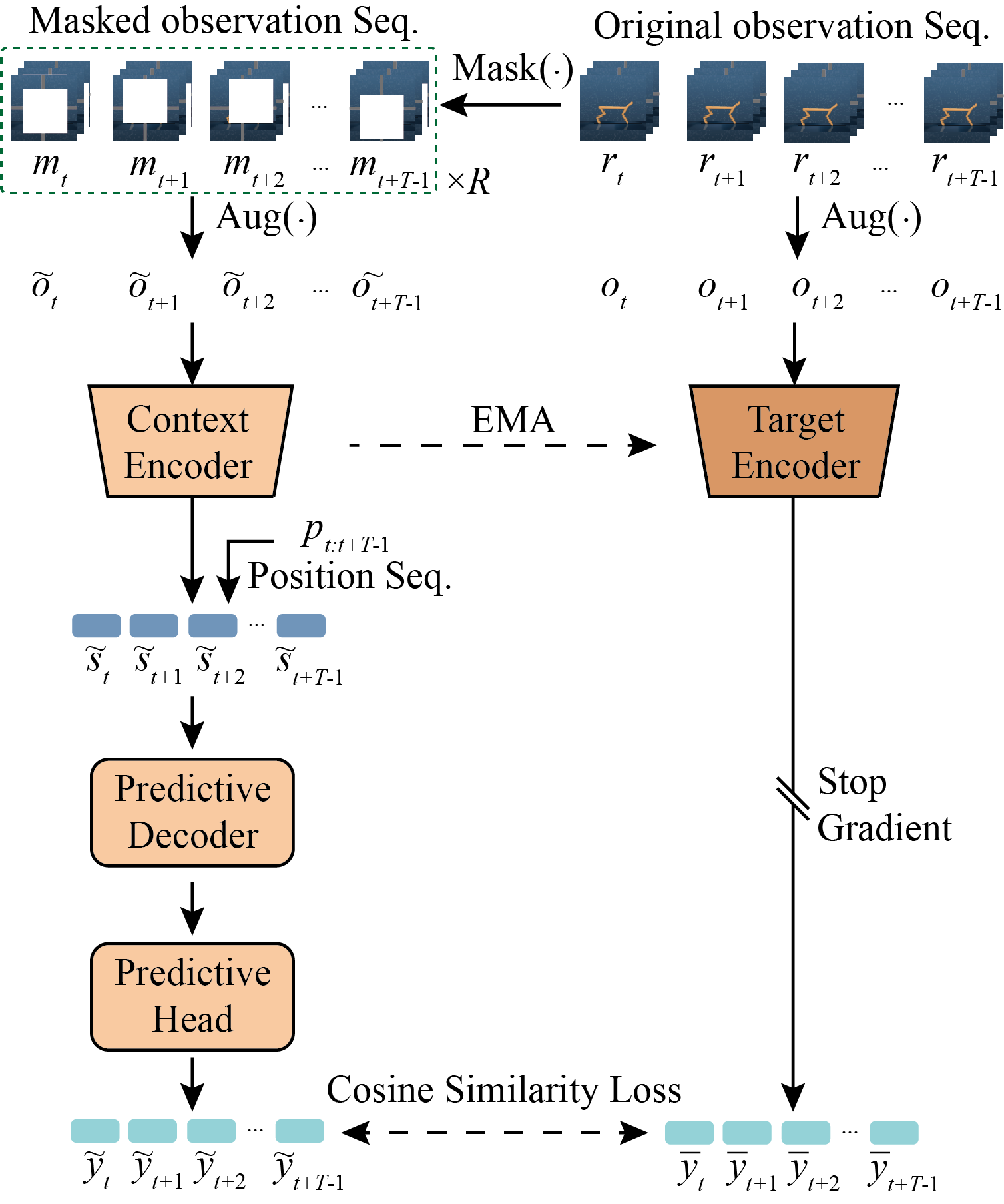}
	\caption{
		The framework of model MPR involves masking the original observation sequence in both spatial and temporal dimensions, followed by encoding the images into feature sequences using an online encoder. We introduce a prediction decoder to predict and reconstruct missing information in the feature sequences augmented with positional information within the latent space. The network is guided to learn effective representations through a similarity-based distance loss function.
		The target encoder weights are the moving average (EMA) of the context encoder weights.
		$R$ represents the number of times the sequence is replicated.
	}
	\label{fig:framework}
\end{figure}

Vision-based reinforcement learning typically employs self-supervised representation learning methods as auxiliary objectives to learn or predict effective state representations. These auxiliary objectives can be combined with different reinforcement learning algorithms, such as Soft Actor-Critic \cite{haarnoja2018soft}, to promote policy learning and enhance sample efficiency.
The foundation of the Mask-based Predictive Architecture (MPR) draws inspiration from joint-embedding predictive architectures, as highlighted in prior works \cite{lecun2022path,assran2023self}. Instead of predicting the features of a state in pixel space, MPR learns to predict them in the embedding space. This approach helps to reduce the interference of redundant information in high-dimensional state-action spaces. The core of MPR lies in predicting the features of target sequences in the latent space through a series of masked images, thereby promoting the understanding of visual signals and tasks by RL agents. We illustrate the framework of MPR in Figure \ref{fig:framework}.

Given a continuous original observation sequence $\mathbf{r}=\left\{r_t,r_{t+1},\cdots, r_{t+T-1}\right\}$, where $T$ represents for sequence length. 
After image augmentation and mask, we obtain a continuous target sequence $\mathbf{o}=\left\{o_t, o_{t+1}, \cdots, o_{t+T-1}\right\}$ and then $\mathbf{o}$ is fed into a target encoder and obtain latent embedding $\bar{\mathbf{y}}=\left\{\bar{y}_t, \bar{y}_{t+1}, \cdots, \bar{y}_{t+T-1}\right\}$ as the target to predict. 
We then divide the original observation sequence $\mathbf{r}$ into different blocks equally and apply the same mask to each image in the same block to form masked observation sequence $\mathbf{m}=\left\{m_t,m_{t+1},\cdots,m_{t+T-1}\right\}$. 
We can determine whether to generate another mask for sequence $\mathbf{m}$ by setting $R$, which provides different perspectives of information.
When $R=2$, the length of $m$ will increase to $2*T$. 
In DMControl experiments, $R=1$, while in Atari benchmarks, $R=2$.

After image augmentation, we obtain the context sequence $\tilde{\mathbf{o}} = \left\{\tilde{o}_t, \tilde{o}_{t+1}, \cdots, \tilde{o}_{t+T-1}\right\}$.
We utilize convolutional networks as both the context encoder and the target encoder, along with a transformer encoder serving as the predictive decoder. The features of the context sequence $\tilde{\mathbf{o}}$ in the latent space are then passed through a projection head to predict the features $\bar{\mathbf{y}}$ of the target sequence $\mathbf{o}$.

\begin{figure}[t]
	\centering
	\includegraphics[scale=0.75]{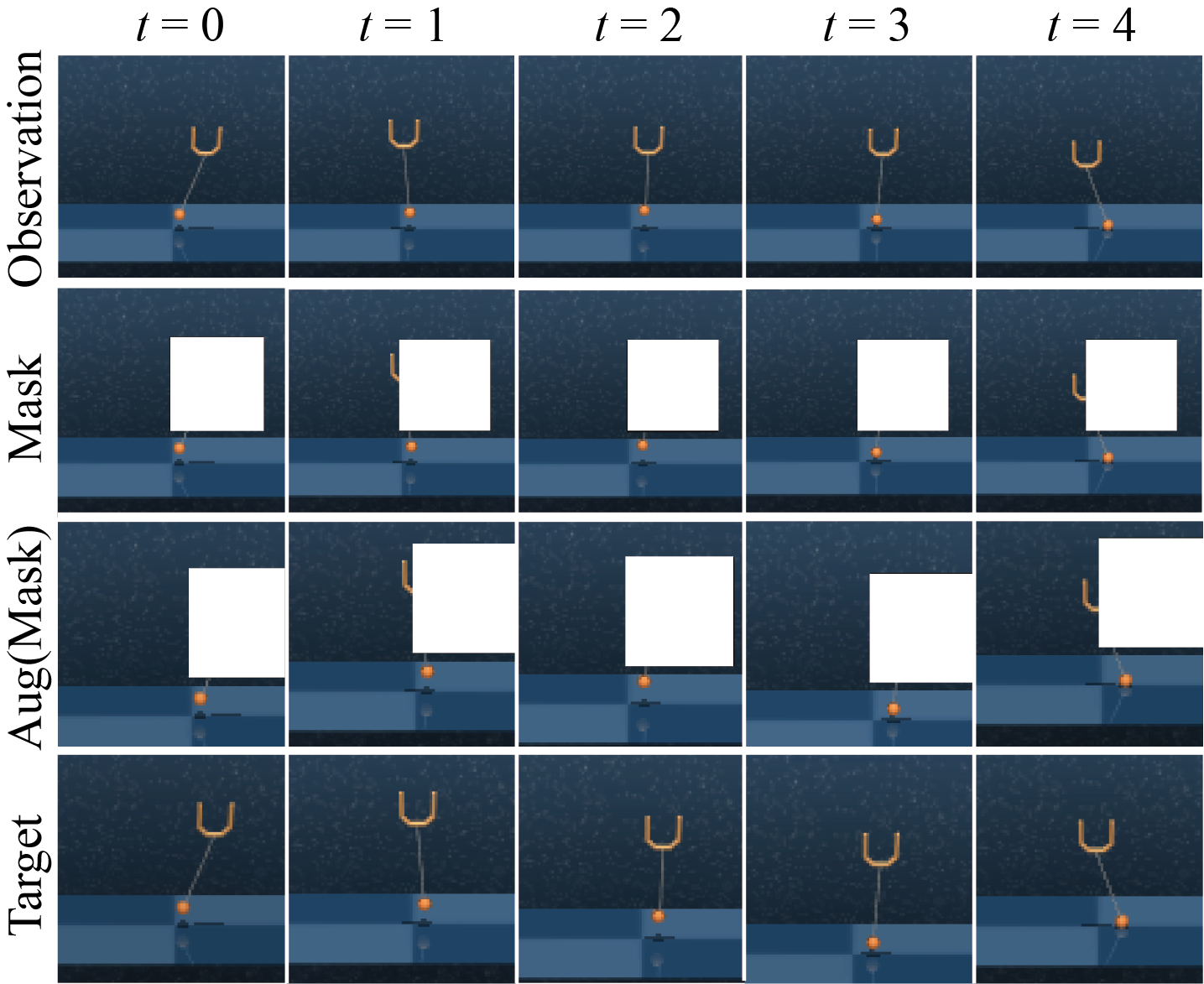}
	\caption{Examples of the original observation sequence, along with their corresponding mask sequences, as well as the context sequence (Aug(Mask)) and target sequence after image augmentation.
	}
	\label{fig:imags}
\end{figure}

\textbf{Context.} 
As described above, our context sequence $\tilde{\mathbf{o}}$ consists of masked images, each with a size of $100\times100$ pixels. We randomly add a $40\times40$  mask to the center of each image. In many reinforcement learning benchmarks, crucial information for tasks is often located at the center of the image. Our aim is to encourage the context encoder $f$ to predict the missing features to provide a better representation for the policy network effectively.
Figure \ref{fig:imags} illustrates the states at five time steps of the original observation sequence, along with their corresponding mask sequences, as well as the context sequence and target sequence after image augmentation.

\begin{table*}[ht]
	
	\footnotesize 
	\centering
	\caption{Comparison results ($mean \pm std$) on the DMControl-100k and DMControl-500k benchmarks.}
	\label{table:dmc_compare}
	\resizebox{\textwidth}{!}{
		\begin{tabular}{l c c c c c c c c c}
			\toprule
			\textbf{100k Step Scores} & \textbf{PlaNet} & \textbf{Dreamer} & \textbf{SAC+AE} & \textbf{CURL} & \textbf{DrQ} & \textbf{PlayVirtual} & \textbf{MLR} &\textbf{MJE}\\ \hline
			\specialrule{0em}{1.5pt}{1pt}   
			Finger, spin & 136 $\pm$ 216  & 341 $\pm$ 70   & 740 $\pm$ 64    & 767 $\pm$ 56  & 901 $\pm$ 104  & {915 $\pm$ 49}  & 907 $\pm$ 58 &\textbf{942$\pm$60}\\
			Cartpole, swingup & 297 $\pm$ 39   & 326 $\pm$ 27   & 311 $\pm$ 11    & 582 $\pm$ 146 &  759 $\pm$ 92 & 816 $\pm$ 36   & 806 $\pm$ 48 & \textbf{863 $\pm$ 15}\\
			Reacher, easy & 20 $\pm$ 50    & 314 $\pm$ 155  & 274 $\pm$ 14    & 538 $\pm$ 233  & 601 $\pm$ 213 & 785 $\pm$ 142   & {866 $\pm$ 103} & \textbf{935$\pm$17}\\
			Cheetah, run & 138 $\pm$ 88   & 235 $\pm$ 137  & 267 $\pm$ 24    & 299 $\pm$ 48  & 344 $\pm$ 67  & 474 $\pm$ 50   & {482 $\pm$ 38}& \textbf{538$\pm$43}\\
			Walker, walk  & 224 $\pm$ 48   & 277 $\pm$ 12   & 394 $\pm$ 22    & 403 $\pm$ 24  & 612 $\pm$ 164  & 460 $\pm$ 173   & 643 $\pm$ 114 &\textbf{774$\pm$136}\\
			Ball in cup, catch & 0 $\pm$ 0      & 246 $\pm$ 174  & 391 $\pm$ 82    & 769 $\pm$ 43  & 913 $\pm$ 53 & 926 $\pm$ 31 & \textbf{933 $\pm$ 16} & 928$\pm$ 35\\ 
			\midrule
			Mean & 135.8 & 289.8 & 396.2 & 559.7 & 688.3 & 729.3 & {772.8}&\textbf{830.0} \\ 
			Median & 137.0 & 295.5 & 351.0 & 560.0 & 685.5 & 800.5 &836.0& \textbf{870.0} \\ 
			\midrule
			\textbf{500k Step Scores}  & & & & & & \\ \midrule
			Finger, spin & 561 $\pm$ 284  & 796 $\pm$ 183  & 884 $\pm$ 128   & 926 $\pm$ 45 & 938 $\pm$ 103 & 963 $\pm$ 40 & {973 $\pm$ 31}&\textbf{983$\pm$2}\\
			Cartpole, swingup & 475 $\pm$ 71   & 762 $\pm$ 27   & 735 $\pm$ 63    & 841 $\pm$ 45& 868 $\pm$ 10 & 865 $\pm$ 11 & \textbf{872 $\pm$ 5}&{859$\pm$2} \\
			Reacher, easy & 210 $\pm$ 390  & 793 $\pm$ 164  & 627 $\pm$ 58    & 929 $\pm$ 44 & 942 $\pm$ 71&942 $\pm$ 66 & {957 $\pm$ 41}&\textbf{971 $\pm$ 21}\\
			Cheetah, run & 305 $\pm$ 131  & 570 $\pm$ 253  & 550 $\pm$ 34    & 518 $\pm$ 28  & 660 $\pm$ 96 & {719 $\pm$ 51}  & 674 $\pm$ 37 & \textbf{730 $\pm$ 3}\\
			Walker, walk  & 351 $\pm$ 58   & 897 $\pm$ 49   & 847 $\pm$ 48    & 902 $\pm$ 43  & 921 $\pm$ 45 &928 $\pm$ 30 & {939 $\pm$ 10}&\textbf{945$\pm$7}\\
			Ball in cup, catch  & 460 $\pm$ 380  & 879 $\pm$ 87   & 794 $\pm$ 58    & 959 $\pm$ 27 & 963 $\pm$ 9 & {967 $\pm$ 5} & 964 $\pm$ 14&\textbf{973 $\pm$ 6}\\ 
			\midrule
			Mean & 393.7 & 782.8 & 739.5 & 845.8 & 882.0 & {897.3} & 896.5 &\textbf{910.1}\\ 
			Median & 405.5 & 794.5 & 764.5 & 914.0 & 929.5 & 935.0 & {948.0}&\textbf{948.5} \\
			\bottomrule
		\end{tabular}
	}
\end{table*}

\textbf{Targets.}
Given an input $\mathbf{o}$, we obtain the high-level representation of each image in \( \bar{\mathbf{y}} \) through the target encoder $\bar{f}$. These high-level features of the images contain contextual information, serving as the learning targets, which can guide the model to acquire more effective representations and enhance the agent's understanding of specific task. 
The parameters of  the target encoder $\theta_{\bar{f}}$ are updated by an exponential moving average (EMA) of the context encoder weights $\theta_{f}$ with the momentum coefficient $\alpha\in[0, 1)$, as below:
\begin{equation}
	\theta_{\bar{f}}\leftarrow \tau\theta_{\bar{f}} + (1-\tau)\theta_{f}
\end{equation}

\textbf{Prediction.}
\citet{assran2023self} predicts target information from the same single image. 
In our implementation, considering the correlation between sequences in reinforcement learning, we not only use single images to predict target vectors, but also leverage the rich contextual information contained within the sequences to assist in predicting missing information in the images.
Therefore, we employ a network consisting of two layers of attention mechanisms, referred to as the predictive decoder, to generate prediction information. Following the concept of mask-based modeling in computer vision, the decoder in our model primarily predicts mask features to fill in the missing information. 
Reconstructing images using predicted features in pixel space is computationally expensive. 
Therefore, similar to the approach in \citet{assran2023self} and \citet{yu2022mask}, we directly predict the target feature $\tilde{\mathbf{y}}$ in the latent space to avoid redundant learning. Additionally, we incorporate learnable position encodings into the feature $\tilde{\mathbf{s}}$. Thus, the inputs of the predictive decoder can be mathematically represented as follows:
\begin{equation}
	\mathbf{x} = \left\{\tilde{s}_{t}+ p_t\right\}
\end{equation}
where $t\in\left\{t, t+1,...,t+T-1\right\}$. the input token sequence is fed into a Transformer encoder consisting of  $l$ attention layers. Each layer is composed of a multi-headed self-attention (MHSA) layer, a layer normalisation (LN), and a multilayer perception (MLP) blocks. The process can be described as follows:
\begin{equation}
	\mathbf{z}^l = MHSA(LN(\mathbf{x}^l)) + \mathbf{x}^l
\end{equation}
\begin{equation}
	\mathbf{x}^{l+1} = MLP(LN(\mathbf{z}^l)) + \mathbf{z}^l
\end{equation}
where $l\in \left\{0,1\right\}$, $\mathbf{x}^0=\mathbf{x}$.

The output sequence $\mathbf{x}^2$ consists of latent representation sequences predicted by the decoder. We employ a projection head to project the output sequence, forming the final predicted latent representation$\left\{\tilde{y}_t, \tilde{y}_{t+1},...,\tilde{y}_{t+T-1}\right\}$. Next, we describe the reconstruction loss.

\textbf{Loss.} Inspired by~\citet{schwarzer2020data, yu2022mask}, 
we directly compute the cosine similarity loss between the predicted latent representation and the target representations. Therefore, the expression for the loss fuction is as follows:
\begin{equation}
	\mathcal{L}_{mpr}=1 - \frac{1}{T}\sum_{i=t}^{t+T-1}\frac{\tilde{y}_i}{\|\bar{y}_{i}\|_{2}}\frac{\bar{y}_i}{\|\bar{y}_{i}\|_{2}}
	\label{eq:ssl loss}
\end{equation}
The self-supervised method proposed in this work serves as an auxiliary task and is optimized together with the policy network. Therefore, the overall loss function is represented as follows:
\begin{equation}
	\mathcal{L}_{total}=\mathcal{L}_{rl}+\lambda\mathcal{L}_{mpr}
\end{equation}
where $\mathcal{L}_{rl}$ and $\mathcal{L}_{mpr}$ respectively denote the base reinforcement learning loss and the self-supervised loss proposed by MPR, with $\lambda$ representing a hyperparameter used to balance the loss functions. Vision-based reinforcement learning models generally consist of two parts: a representation learning network and a policy network. The target encoder of our proposed model will feed well-learned visual representations into the RL model as low-dimensional state inputs for reinforcement learning. In the testing phase, the context encoder and predictive decoder cease operation, leaving only the trained momentum target encoder in use.
Please refer to the Appendix \ref{Implementation Details of DMContol} for more details on positional embeddings and algorithmic details, including pseudocode.

\begin{table*}[ht]
	\footnotesize 
	\centering
	\caption{Comparison on the Atari-100k benchmark. The MPR results are averaged across 10 seeds, each seed for 100 episodes; the results of SimPLe, CURL, DrQ, SPR, MLR and CoIT are taken from their report results.} 
	\label{table:atari_compare}
			\begin{tabular}{l c cc c c c c c c}
				\toprule
				\textbf{Game}                                  & \textbf{Human} & \textbf{Random} &\textbf{SimPLe} & \textbf{CURL} & \textbf{DrQ} & \textbf{SPR}     & \textbf{MLR}    & \textbf{CoIT}   & \textbf{MPR}       \\ 
				\midrule
				Alien 													& 7127.7         & 227.8                            &616.9 & 558.2           & 771.2                & 801.5            		 & 990.1           					& \textbf{1206.7} 		& 1186.0             \\
				Amidar                                              & 1719.5         & 5.8                                 &88.0 & 142.1             & 102.8                 & 176.3           	    & 227.7           					& 182.3           				& \textbf{254.1}     \\
				Assault                                              & 742.0          & 222.4                             &527.2 & 600.6            & 452.4              & 571.0            		  & 643.7           					& 635.7           			& \textbf{667.3}     \\
				Asterix                                              & 8503.3         & 210.0                            &\textbf{1128.3} & 734.5             & 603.5             & {977.8}   	& 883.7           					& 709.0           				& 799.1             \\
				Bank Heist                                       & 753.1          & 14.2                                 &34.2 & 131.6               & 168.9              & \textbf{380.9}   	& 180.3           					& 124.9           				& 67.3               \\
				Battle Zone                                     & 37187.5        & 2360.0                         &5184.4 & 14870.0         & 12954.0          & 16651.0          		& {16080.0}       				& 13760.0         			& \textbf{20274.8}  \\
				Boxing                                             & 12.1           & 0.1                                     &9.1 & 1.2                   & 6.0                  & \textbf{35.8}    		& 26.4            					& 23.6            				& 21.0               \\
				Breakout                                         & 30.5           & 1.7                                    &16.4 & 4.9                  & 16.1                & {17.1}    		   & 16.8            					& 16.1            				& \textbf{18.1}              \\
				Chopper Cmd                                & 7387.8         & 811.0                             &1246.9 & 1058.5           & 780.3              & {974.8}          		   & 910.7           			        & \textbf{1338.0}	 & 837.8              \\
				Crazy Climber                                & 35829.4        & 10780.5                      & \textbf{62583.6} & 12146.5         & 20516.5          &{42923.6}    & 24633.3         			     & 17538.0         			& 24902.36          \\
				Demon Attack                               & 1971.0         & 152.1                                &208.1 & 817.6              & 1113.4             & 545.2            			 & 854.6           			       & 864.6           			& \textbf{1176.7}    \\
				Freeway                                         & 29.6           & 0.0                                    &20.3 & 26.7               & 9.8                   & 24.4             			    & \textbf{30.2}   		       & 29.6            			& 29.6               \\
				Frostbite                                        & 4334.7         & 65.2                                &254.7 & 1181.3            & 331.1              & 1821.5           		      & {2381.1}        		        & 2069.8          			& \textbf{2964.0}   \\
				Gopher                                          & 2412.5         & 257.6                               &771.0 & 669.3            & 636.3              & 715.2            		        & \textbf{822.3}  		     & 746.8           			& 558.8             \\
				Hero                                              & 30826.4        & 1027.0          				  &2656.6 & 6279.3         & 3736.3            & 7019.2           		     &7919.3          			     & 7572.8          			& \textbf{11044.1}  \\
				Jamesbond                                  & 302.8          & 29.0            				       &125.3 & \textbf{471.0}            & 236.0               & 349.0            		       & {423.2}  		   & 336.0           			& 400.1              \\
				Kangaroo                                     & 3035.0         & 52.0            				    &323.1 & 872.5            & 940.6              & 3276.4           		     &\textbf{8516.0} 			& 4116.6          			& 1286.3             \\
				Krull                                              & 2665.5         & 1598.0          				    &4539.9 & 4229.6         & {4018.1}          & 3688.9          		      &3923.1          				   & 3436.2          			& \textbf{4724.5}    \\
				Kung Fu Master                          & 22736.3        & 258.5           				     &17257.2 & 14307.8        & 9111.0            & {13192.7}        	       &10652.0         			   & 9250.0          			& \textbf{17797.4}  \\
				Ms Pacman                                & 6951.6         & 307.3           				      &1480.0 & 1465.5          & 960.5              & 1313.2           		    &{1481.3}        			    & \textbf{1509.6} 		& 1413.8             \\
				Pong                                            & 14.6           & -20.7           				           &12.8 & -16.5             & -8.5                & -5.9             		        &\textbf{4.9}    			  & 1.5             				& -6.9              \\
				Private Eye                                 & 69571.3        & 24.9            				        &58.3 & \textbf{218.4}             & -13.6              & 124.0             	         &100.0               			  & {145.7}  		& 100.0              \\
				Qbert                                          & 13455.0        & 163.9          				      &1288.8 & 1042.4           & 854.4            & 669.1            	         &3410.4          		       	& 2117.5          				& \textbf{4333.0}    \\
				Road Runner                             & 7845.0         & 11.5            					      &5640.6 & 5661.0            & 8895.1           & 14220.5          	        &12049.7         			  & 11758.5         			& \textbf{16157.8}   \\
				Seaquest                                   & 42054.7        & 68.4            				       &\textbf{683.3} & 384.5               & 301.2           & 583.1            		       &{628.3}         			    & 554.0           			& {677.2}     \\
				Up N Down                               & 11693.2        & 533.4           				        &3350.3 & 2955.2              & 3180.8       & \textbf{28138.5}       &6675.7          			    & 4734.2          			& 8630.4             \\ 
				\bottomrule& 
			\end{tabular}
		\end{table*}

\begin{figure*}[ht]
	\centering    
	\begin{subfigure}{0.32\textwidth}
		\centering
		\includegraphics[scale=0.38]{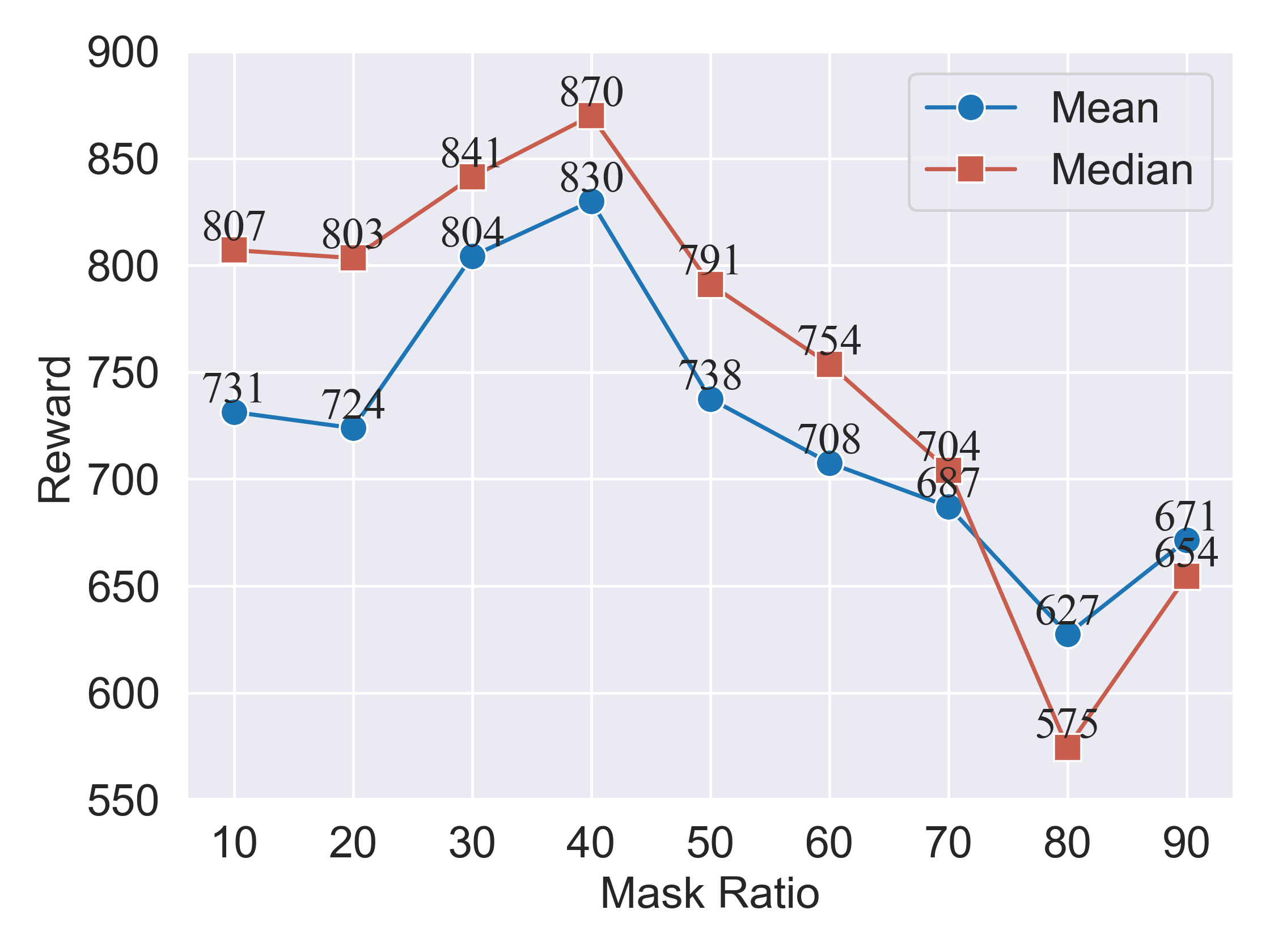}
		\caption{}
		\label{fig:ablation mask ratio}
	\end{subfigure}
	\hfill
	\begin{subfigure}{0.32\textwidth}
		\centering
		\includegraphics[scale=0.38]{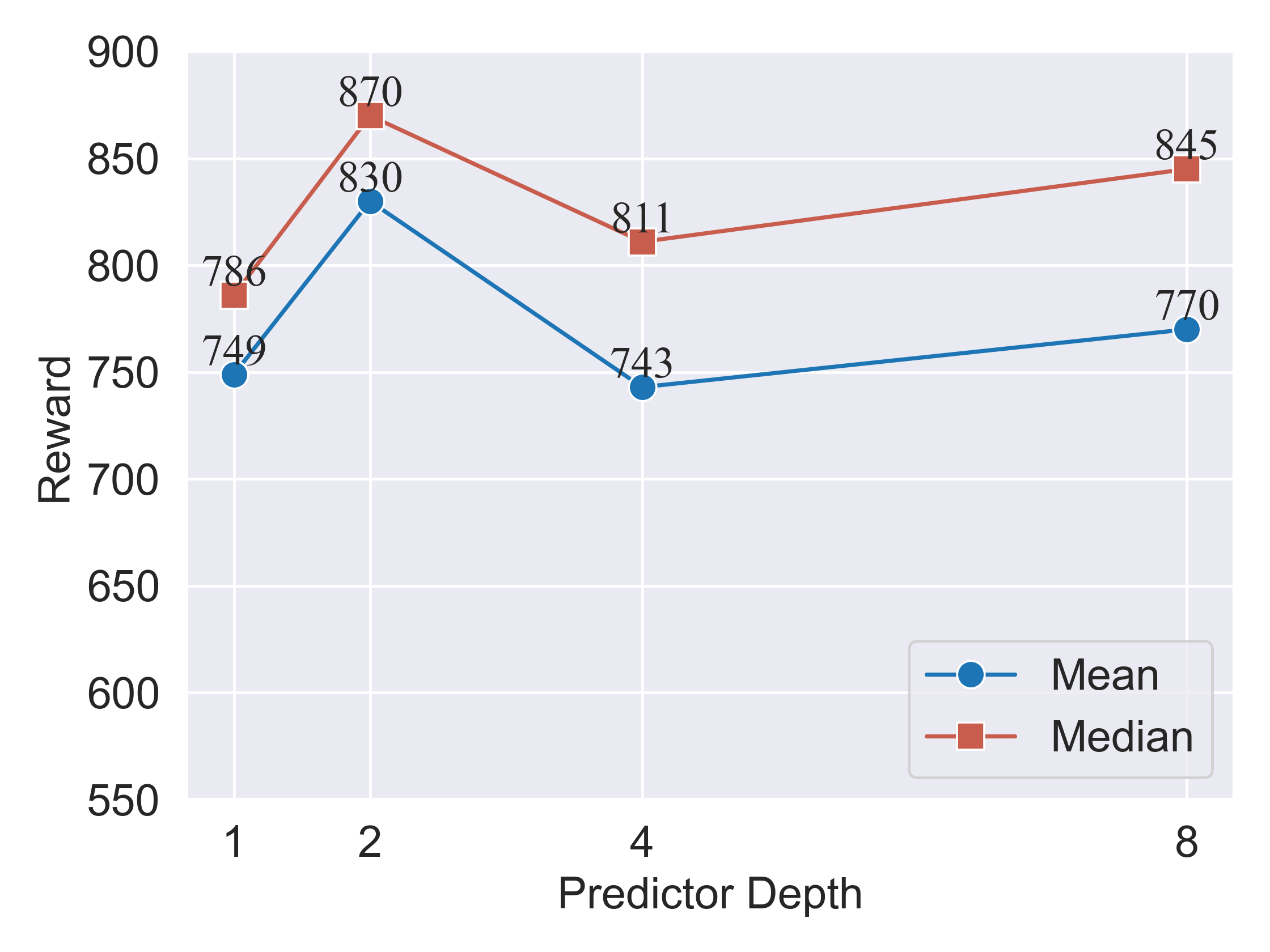}
		\caption{}
		\label{fig:ablation predictor depth}
	\end{subfigure}
	\hfill
	\begin{subfigure}{0.32\textwidth}
		\centering
		\includegraphics[scale=0.38]{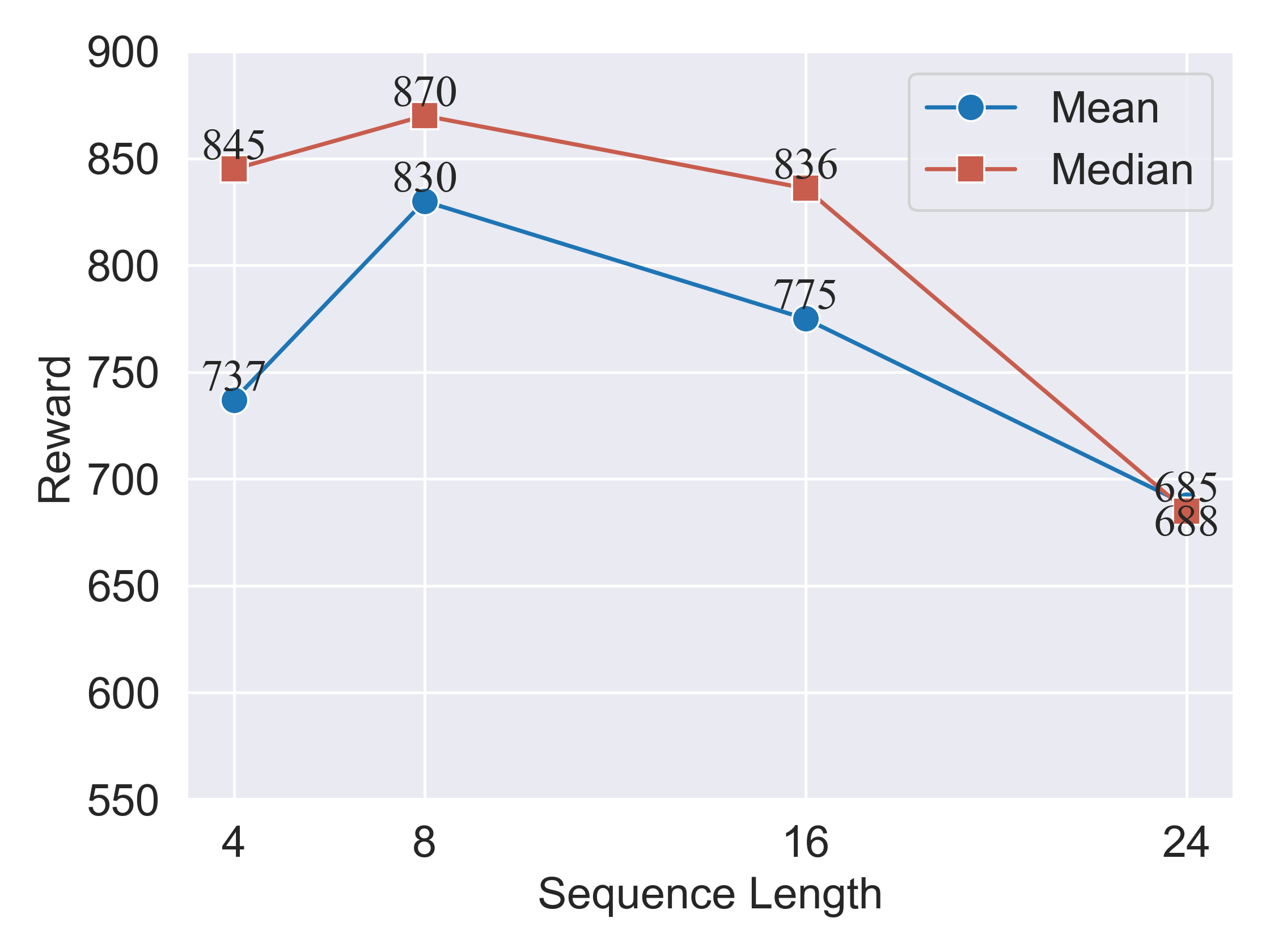}
		\caption{}
		\label{fig:ablation sequence length}
	\end{subfigure}
	\caption{Ablation experiments of mask ratio, predictor depth and sequence length. The result of each model is averaged over 3 random seeds.}
	\label{fig:ablation mask ratio predictor depth and sequence length}
\end{figure*}

\section{Experiment}
\subsection{Setup}
\textbf{Environments and Evaluation.}
We evaluated the sample efficiency of our proposed model on both the DeepMind Control Suite (DMControl) \cite{tassa2018deepmind}, representing continuous action spaces, and Atari \cite{bellemare2013arcade}, representing discrete action spaces. On the DMControl benchmark, align with previous works \cite{laskin2020reinforcement, laskin2020curl, yarats2020image, yu2022mask}, we selected 6 tasks: \textit{Finger, spin; Cartpole, swingup; Reacher, easy; Cheetah, run; Walker, walk; and Ball in cup, catch}.
The score range for each environment is from 0 to 1000 \cite{tassa2018deepmind}. 
For continuous control,
We conducted evaluations with 10 episodes each at 100k and 500k environment steps, denoted as DMControl-100k and DMControl-500k benchmarks, respectively. Similar to representation learning, we select the best-performing policy during training for evaluation.
For discrete control, we tested our model on the Atari-100k benchmark, which includes 26 Atari games. In each environment, the model interacted for 100k steps, with each interaction repeating actions 4 times, resulting in a total of 400k interactions. 
Considering the high variance in scores on this benchmark, we conducted tests under 10 different seeds, each run with 100 episodes \cite{schwarzer2020data}. 

\textbf{Implementation.}
In continuous control experiments, we employ SAC \cite{haarnoja2018soft} as the baseline agent. We randomly sample masks of size 40$\times$40 and overlay them onto simulation-rendered images sized 100$\times$100. Following image augmentation, the images are resized to 84$\times$84.
For discrete control, we employ Rainbow \cite{hessel2018rainbow}  and MLR \cite{yu2022mask} as the baseline agent. The size of Atari-rendered images is 84$\times$84. To maintain a masking rate of approximately 40\%, the size of masks randomly sampled in Atari games is 32 $\times$32.
In the loss function, the coefficient $\lambda$ adjusts the balance between $\mathcal{L}_{rl}$ and $\mathcal{L}_{mpr}$, and we set $\lambda$ to 1. For the length of sequences, we uniformly set $T$ to 8.
The block size is related to the masking policy, and specific settings for DMcontrol tasks will be discussed in detail in the ablation experiments. The block size in Atari games is uniformly set to 8.

\begin{table*}[h]
	\centering
	\caption{Ablation experiments on block size and mask strategy.
		When the block size is set to 1, we add a unique mask for each state, denoted as MPR\textsubscript{spatial}. When the block size is set to 8, all states use the same mask, denoted as MPR\textsubscript{time}. For other block size settings, corresponding spatial-temporal masking strategies are applied.
	}
	\begin{tabular}{lcccc}
		\toprule
		Environment        &      1      &      2      &      4      &      8      \\ \midrule
		Finger, spin       & 559$\pm$89  & 869$\pm$78  & 620$\pm$45  & \textbf{942$\pm$60} \\
		Cartpole, swingup  & 828$\pm$25  & 847$\pm$35  & \textbf{863$\pm$15}  & 797$\pm$33  \\
		Reacher, easy      & 933$\pm$55  & \textbf{935$\pm$17}  &  916$\pm$53  & 841$\pm$50  \\
		Cheetah, run       &\textbf{538$\pm$43}  & 508$\pm$15  &  500$\pm$2  & 486$\pm$37  \\
		Walker, walk       & 767$\pm$30  & 577$\pm$163 & 427$\pm$75  & \textbf{774$\pm$136} \\
		Ball in cup, catch & 878$\pm$107 & 886$\pm$ 52 & 875$\pm$120 & \textbf{928$\pm$35}  \\ \midrule
		Mean               &     751     &     770     &     700     &     795     \\
		Median             &     788     &     862     &     789     &     813     \\ \bottomrule
	\end{tabular}
	\label{table:mask strategy and block size}
\end{table*}

\subsection{Comparison with State-of-the-Arts}
\textbf{DMControl.}
We compare our proposed method with previous sample efficiency models, including PlaNet \cite{hafner2019learning}, Dreamer \cite{hafner2019dream}, SLAC \cite{lee2020stochastic}, CURL \cite{laskin2020curl}, DrQ \cite{yarats2020image}, and MLR \cite{yu2022mask}. The comparison results are shown in Table \ref{table:dmc_compare}, where we conduct experiments with 10 different random seeds, each seed with 10 episodes, and the displayed results are presented as $mean\pm std$. 
Since our model is similar to MLR, we mainly compare with MLR:
In the DMControl 100k experiment, it can be observed that:

(\romannumeral1) MPR achieved performance improvement in 5/6 tasks, with an average task score of 830.0, which is a 7.5\% increase compared to MLR (772.8), with a median score improvement of 4.1\%.

(\romannumeral2) Considering the MPR model has fewer network parameters and higher performance, it can be argued that our proposed masking strategy effectively improves upon the MLR model, achieving higher sample efficiency.

(\romannumeral3) In the DMControl 500k experiment, our model outperforms the previous model in 5 of 6 tasks. Compared to MLR \cite{yu2022mask}, the average task performance improved by 1.5\%. The MPR and MLR perform similarly on DMControl-500k, but MPR requires less computational resources and achieves higher average scores.

\textbf{Atari-100k.}
To validate the effectiveness of the masking policy proposed in this paper, we directly apply the masking policy to MLR and set $R=2$.
We compare our proposed model with the state-of-the-art (SOTA) model-free methods for discrete control, including SimPLe \cite{kaiser2019model}, 
CURL \cite{laskin2020curl}, DrQ \cite{yarats2020image}, SPR \cite{schwarzer2020data}, MLR \cite{yu2022mask} and CoIT \cite{liu2022data}. Our proposed model MPR for Atari is based on Rainbow \cite{hessel2018rainbow} and MLR \cite{yu2022mask}.
We select the best-performing policy saved within 100k steps to evaluate the performance of MPR in Atari games.
We conduct experiments with 10 randomly chosen seeds, each run comprising 100 episodes. We average MPR's performance over the 10 random seeds.
The results for Atari 100k are shown in Table \ref{table:atari_compare}. Below are the key findings:
(\romannumeral1) MPR outperforms other models in 11 out of 26 games in the Atari benchmark. In the remaining games, MPR's scores are also competitive.
(\romannumeral2) MPR achieves scores exceeding human performance in 5 games (\textit{Boxing}, \textit{Freeway}, \textit{Jamesbond}, \textit{Krull}, \textit{Road Runner}).
(\romannumeral3) The masking policy and the setting of $R$ in MPR are effective compared to MLR. In discrete control tasks, where there is significant discontinuity between adjacent states, the incorporation of multi-view information benefits MPR by facilitating smoother learning, thereby enhancing its policy.

\subsection{Ablation Study}


To demonstrate the effectiveness of the auxiliary task proposed in this work, in this section, we conduct ablation experiments on the DMControl-100k benchmark. We systematically investigate the effects of mask ratio, block size and mask strategy, predictive decoder depth, sequence length and alternative design. We evaluate each model with 3 different seeds.

\textbf{Mask Ratio.}
Previous works on masked image modeling \cite{he2022masked, xie2022simmim, yu2022mask} reveal that the mask ratio has a significant impact on model performance. Consistent with previous studies \cite{yu2022mask}, we investigated the effect of different proportions of masks on model performance. With intervals of 10\%, we gradually increased the mask ratio from 10\% to 90\% to observe the variation in model performance. The experimental results are shown in Figure \ref{fig:ablation mask ratio}. It can be observed that the performance of MPR initially improves and then declines with the increasing mask ratio. The performance of MPR is optimal when the mask ratio is 40\%.

\textbf{Block Size and Mask Strategy.}
Previous studies \cite{yu2022mask, he2022masked} have shown that the mask strategy has a significant impact on representation learning. In MPR, the mask strategy is related to the block size. Adopting the same mask strategy in the temporal dimension is denoted as MPR\textsubscript{time}, corresponding to a block size of 8, where each image at every time step uses the same mask. Masking the context sequence in the spatial dimension is denoted as MPR\textsubscript{spatial}, corresponding to a block size of 1, where each image at every time step uses a different mask.
When the block size is set to 2 and 4, it corresponds to the spatial-temporal mask strategy.
We conduct experiments with block sizes set to 1, 2, 4, and 8, and the experimental results are shown in Table \ref{table:mask strategy and block size}. Different tasks require different mask strategies. The \textit{Finger-spin}, \textit{Walker-walk}, and \textit{Ball in cup-catch} tasks tend to favor the time mask strategy (block size: 8). The \textit{Cheetah-run} task performs best under the spatial mask strategy (block size: 1), while the \textit{Cartpole-swingup} (block size: 4) and \textit{reacher-easy} (block size: 2) tasks prefer the spatial-temporal mask strategy.
We report the best scores achieved by MPR under their respective optimal block sizes.

\textbf{Predictor depth.}
We aim for the model to learn good representations predominantly from the encoder, as these representations are directly used for policy learning. We investigated the impact of the predictive decoder depth on model performance. We set the predictive decoder depth to 1, 2 (MPR), 4, and 8, respectively. Generally, a deeper predictive decoder depth implies lower sample efficiency and more wall clock time. The results are shown in Figure \ref{fig:ablation predictor depth}. MPR achieves optimal performance when the depth is 2; increasing the predictive decoder depth reduces the effectiveness of the model's policy learning. 
 For more detailed tabular results please refer to Appendix Table \ref{table:ablation on predictor depth}.

\textbf{Sequence Length.}
The sequence length determines the amount of information that can be obtained by MPR. We set the sequence length to 4, 8 (MPR), 16, and 24, respectively, to study the change in model performance. The results are shown in Figure \ref{fig:ablation sequence length}. It can be seen that a smaller sequence length means that the model cannot use contextual information, resulting in an average score of 737. As the sequence length increases, the model's performance improves to 830. Further increasing the sequence length leads to a significant decline in model performance because longer sequences can provide more contextual information, and MPR tends to utilize the predictive decoder to predict latent embeddings by learning from other states.  For more detailed tabular results please refer to Appendix Table \ref{table:ablation on sequence length}.

\begin{figure}[ht]
	\centering
	\includegraphics[scale=0.38]{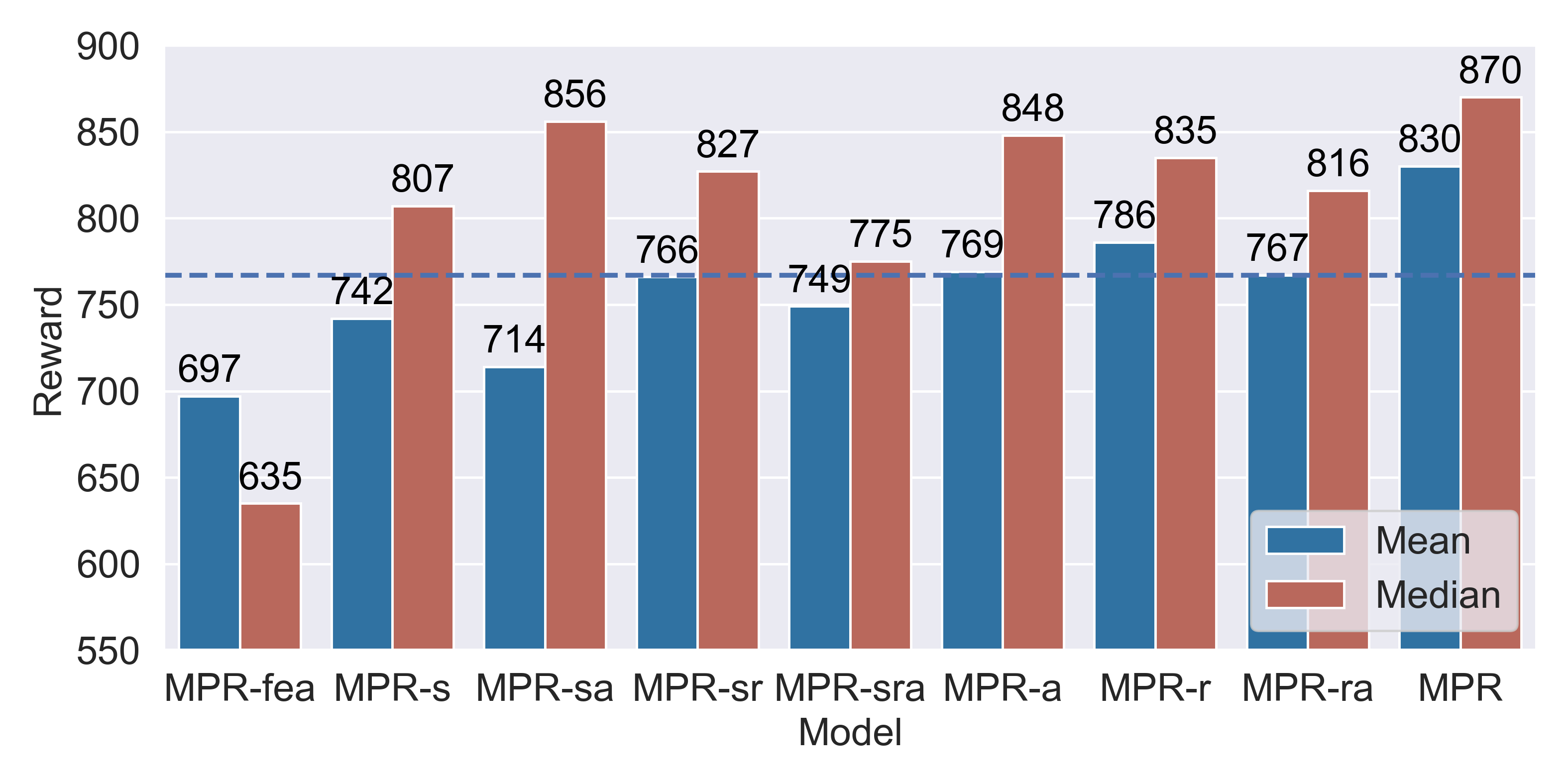}
	\caption{Ablation experiments on action token, $R$ , mean squared loss, and masking features. 
	}
	\label{fig:alternative design}
\end{figure}

\textbf{Alternative Design.}
Lastly, the performance of some alternative designs in MPR on the DMControl benchmark is investigated.
In \cite{yu2022mask}'s work, providing action as sequential information to the latent embedding $\tilde{\mathbf{s}}$ leads to performance improvement. We provide the action sequence to MPR, denoted as MPR-a.
We set $R=2$ to observe the effect on task performance, denoted as MPR-r.
We add action sequence information to MPR-r, denoted as MPR-ra.
Next, we explore the impact of mean squared loss on task performance, denoted as MPR-s.
We add action token information to MPR-s, denoted as MPR-sa.
Following the approach of MPR-r, we form MPR-sr and then add action sequence information, denoted as MPR-sra.
Finally, we investigate the way of masking features, consistent with the mask strategy proposed in our work. We directly mask 40\% of image features, denoted as MPR-fea.
The results are presented in Figure \ref{fig:alternative design}.

 (\romannumeral1) Among all models with cosine similarity as the loss function, the lowest average score for DMControl tasks is 767, indicated by the blue dashed line. It can be observed that all models with mean squared error as the loss function have average task scores lower than 767, indicating that in MPR or reinforcement learning with similar auxiliary prediction targets, it is unnecessary to precisely reconstruct embeddings in the latent space. Several MPR variants with cosine similarity as the loss function are close to ({MPR-a}: 769) or surpass ({MPR-r}: 786) the average task score of {MLR} (772.8), demonstrating the effectiveness of MPR.

 (\romannumeral2) We found that adding action token to MPR or setting $R=2$ both lead to a decrease in MPR performance. This may be because excessively long context sequences fail to provide effective information for the agent and instead reduce the efficiency of representation learning.
%
The detailed table results of all experiments are shown in Appendix Table \ref{table:alternative designs}.

\textbf{Harder Tasks.} 
Finally, we tested our model on more challenging tasks, specifically the \textit{Reacher, hard} and \textit{Walk, run} tasks. We did not search for optimal parameters, but directly set the block size to 4 and $T=8$. We compared our model's performance with that of MLR \cite{yu2022mask}, and the results are presented in Table \ref{tab:results on more challenage tasks}. Our model performs similarly to MLR on both tasks, but excels in the \textit{Reacher, hard} 500k task, with an average score improvement of 130, demonstrating the effectiveness of MLR.

\begin{table}[ht]
	\centering
	\caption{Comparison of MPR and MLR on more challenging DMControl tasks..}
	\label{tab:results on more challenage tasks}
			{\begin{tabular}{cccc}
				\toprule
				            Steps             & Model & Reacher, hard & Walker, run \\ \midrule
				\multirow{2}{*}[-0.0ex]{100k} &  MLR  &  \textbf{624 $\pm$ 220}  & 181 $\pm$ 19  \\
				                                                   &  MPR  &  617 $\pm$ 145   &\textbf{ 184 $\pm$ 25}   \\ \midrule
				\multirow{2}{*}[-0.0ex]{500k} &  MLR  &  844 $\pm$ 129  & \textbf{576 $\pm$ 25}  \\
				                                                    &  MPR  &  \textbf{974 $\pm$ 4}      &571 $\pm$ 8      \\ \bottomrule
			\end{tabular}}
		\end{table}

\section{Discussion}



Based on the concept of joint embedding architecture, we propose MRA, and experimental results on two benchmarks, DMControl and Atari games, demonstrate that our proposed model enhances the sample efficiency of reinforcement learning agents.  However, our work also has some limitations.  Our masking method is randomly sampled, similar to MLR \cite{yu2022mask}, and thus requires tuning of multiple hyperparameters.

Our work shares similarities with MLR \cite{yu2022mask}, with the first difference being the masking method. We utilize a masking method that is independent of patch size, allowing the predictor to decouple from patch size and masking method. The second difference is that we do not incorporate action sequence information in our implementation, as it has been demonstrated not to enhance MPR performance in contimuous control.
The third difference is that we employ only one projection predictor, and there is no online projection head following the predictive decoder. In comparison to MLR, this reduction in the model's parameter count promotes policy learning, rendering it an enhanced version of MLR.

Our work also opens up more possibilities for vision-based RL.  Since the mask is independent of patch size, we can adopt a variety of masks to enhance the effectiveness of self-supervised learning, thereby improving the sample efficiency of RL agents.  For example, SAM \cite{kirillov2023segment} segmentation can be used to generate instance masks or BLIP \cite{li2022blip} can be used to generate language-image masks.  We have demonstrated the effectiveness of our approach, leaving further improvements for future work.

\section{Conclusion}
In this work, 
we investigated the effectiveness of mask modeling in representation learning-based reinforcement learning.  Building upon the concept of joint embedding, we introduced MPR, aiming to enhance policy learning through effective state representations.  Our experimentation on DMcontrol demonstrated the higher sample efficiency of our model.  Additionally, by directly applying the masking strategy to all 26 Atari games, we observed significant improvements in experimental results.  Results from ablation studies further validate the effectiveness of our current model.  The masking concept employed in our model suggests promising avenues for future research to advance the field of sample-efficient reinforcement learning.

\clearpage
\bibliography{aaai23}

\clearpage
\appendix
\onecolumn
\setcounter{section}{0}
\renewcommand{\thesection}{\Alph{section}}

\section{Appendix}
\section{Implementation Details of DMContol}
\label{Implementation Details of DMContol}

In our implementation, our model primarily consists of auxiliary networks and base networks. The base network includes a representation learning network \( f \), parameterized by \( \theta_{{f}} \), and a policy network \( \phi \). The base network \( f \) comprises 4 convolutional layers, each followed by a ReLU activation function, a fully connected layer, and a normalization layer. The policy network \( \phi \) is composed of a multi-layer perceptron.

Our auxiliary network adopts a self-supervised learning architecture. The online network includes an encoder network, a predictive decoder network, and a prediction head network, represented by \( \theta_{f} \), \( \theta_{g} \), and \( \theta_{h} \) respectively. Parameters between \( \theta_{f} \) and \( \theta_{\bar{f}} \) are shared through momentum updates. \( \theta_{g} \) represents a standard self-attention layer with a depth of 2 and the number of attention head was set to 1. 
The predictive decoder in MPR treats the features at each time step as independent tokens; therefore, we need to add positional embeddings for each token. We use sine and cosine functions to build the positional embeddings following \cite{vaswani2017attention}:
\begin{equation}
	p_{(pos,2j)}=sin(pos/10000^{2j/d})
\end{equation}
\begin{equation}
	p_{(pos,2j+1)}=cos(pos/10000^{2j/d})
\end{equation}

The pseudocode description of MPR is as follows:

\begin{algorithm*}[h]
	\renewcommand{\algorithmicrequire}{\textbf{Input:}}
	\renewcommand{\algorithmicensure}{\textbf{Output:}}
	\caption{Training algorithm for MPR.}
	\label{alg1}
	\begin{algorithmic}[1]
		\REQUIRE An online encoder $f$, a momentum encoder $\bar{f}$, a predictive decoder $g$ and a prediction head $h$, a policy network $\phi$, parameterized by $\theta_{f}, \theta_{\bar{f}}, \theta_{g}, \theta_{h}$ and $\theta_{\phi}$, set $R=1$.
		\WHILE{$ train $}
		\STATE Interact with envirpnment and collect the transition: $\mathcal{B}\leftarrow \mathcal{B}\cup(o_t, a_t, o_{t+1}, r)$.
		\STATE Sample a trajectory of $T$ timesteps $\left\{o_t,a_t,o_{t+1}, a_{t+1},\cdots, o_{t+T-1}, a_{t+T-1}\right\}$ from $\mathcal{B}$.
		\STATE Mask the sequence randomly: $\left\{\tilde{o}_t, \tilde{o}_{t+1}, \cdots, \tilde{o}_{t+T-1} \right\} \leftarrow Mask(\left\{o_{t}, o_{t+1},\cdots,o_{t+T-1}\right\})$.
		\STATE Augmentation and then encoding: $\left\{\tilde{s}_t, \tilde{s}_{t+1}, \cdots, \tilde{s}_{t+T-1}\right\}\leftarrow\left\{f(Aug(\tilde{o}_t)), f(Aug(\tilde{o}_{t+1})),\cdots, f(Aug(\tilde{o}_{t+T-1}))\right\}$.
		\STATE Predictive decoding: $ \left\{\hat{s}_t, \hat{s}_{t+1}, \cdots, \hat{s}_{t+T-1}\right\}\leftarrow g(\left\{\tilde{s}_t, \tilde{s}_{t+1}, \cdots, \tilde{s}_{t+T-1}\right\})$
		\STATE Prediction: $\left\{\tilde{y}_t, \tilde{y}_{t+1}, \cdots, \tilde{y}_{t+T-1}\right\}\ \leftarrow h(\left\{\hat{s}_t, \hat{s}_{t+1}, \cdots, \hat{s}_{t+T-1}\right\})$
		\STATE Calculate MPR loss: $\mathcal{L}_{mpr}\leftarrow 1 - \frac{1}{T}\sum_{i=t}^{t+T-1}\frac{\tilde{y}_i}{\|\bar{y}_{i}\|_{2}}\frac{\bar{y}_i}{\|\bar{y}_{i}\|_{2}}$.
		\STATE Calculate RL loss: $\mathcal{L}_{rl}$
		\STATE Calculate total loss: $\mathcal{L}_{total}\leftarrow\mathcal{L}_{rl}+\lambda \mathcal{L}_{mpr}$.
		\STATE Update online parameters: $\theta_{f}, \theta_{g}, \theta_{h}, \theta_{\phi}\leftarrow Optimize(\mathcal{L}_{total})$.
		\STATE Update momentum parameters: $\theta_{\bar{f}}\leftarrow \tau\theta_{\bar{f}} + (1-\tau)\theta_{f}$. 
		\ENDWHILE
	\end{algorithmic}  
\end{algorithm*}

\clearpage
\section{Implementation Details of Atari 100k}

\begin{figure*}[h]
	\centering
	\includegraphics[scale=0.95]{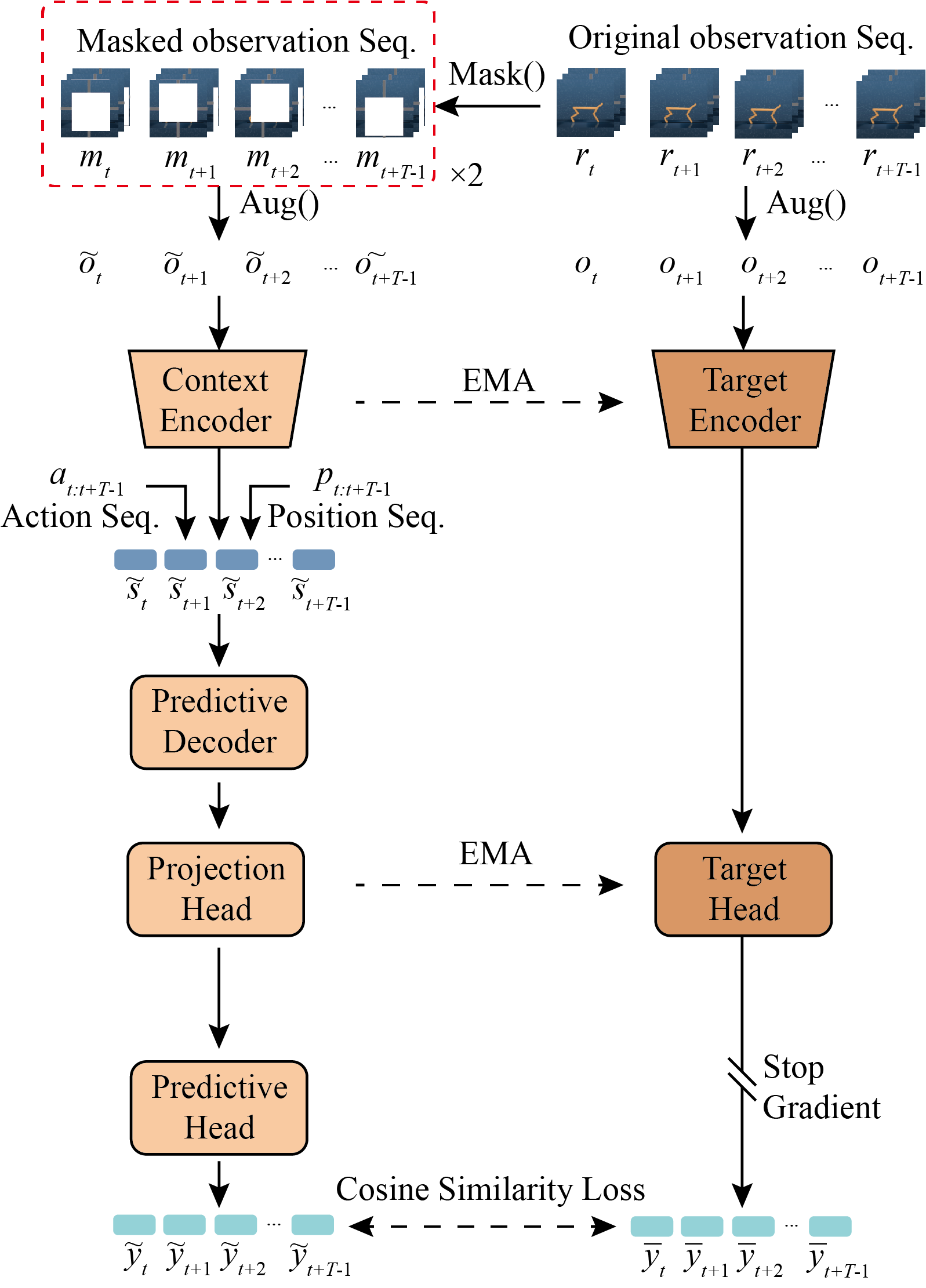}
	\caption{
		The framework of MPR on Atari games.
	}
	\label{fig:framework_atari}
\end{figure*}

Due to the effectiveness and flexibility of the MPR framework, we incorporate some key ideas from this model into discrete control settings. In DMControl, we choose SAC as the policy network for continuous control. For Atari 100k, we select MLR \cite{yu2022mask} (\url{https://github.com/microsoft/Mask-based-Latent-Reconstruction}) as our baseline code, with the mask policy and $R=2$, without any other modifications. We illustrate the modifications of MPR to MLR in the red dashed box in Figure \ref{fig:framework_atari}.
Given the high variance nature of the Atari100k benchmark, we evaluate using 10 random seeds, running over 100 episodes for each random seed, and report the average score for each game.

\clearpage
\section{More Experimental Reaults}
\subsection{Table results  in continuous control}
Table \ref{table:ablation on predictor depth} presents the impact of predictor depth on model performance in the ablation experiments of continuous control. 

\begin{table*}[h]
	\centering
	\caption{Ablation experiments of predictive decoder depth. We report the specific results on DMControl-100k based on 3 random seeds.}
	\begin{tabular}{lcccc}
		\toprule
		Environment        & 1 & 2 (MPR) & 4 & 8 \\ \midrule
		Finger, spin       &785$\pm$262   & 942$\pm$60                        &782$\pm$284   &859$\pm$28   \\
		Cartpole, swingup  &864$\pm$2   &863$\pm$15                         &848$\pm$11   &863$\pm$10   \\
		Reacher, easy      &845$\pm$186   &935$\pm$17                         &844$\pm$88   &780$\pm$132   \\
		Cheetah, run       &463$\pm$116   &538$\pm$43                         &506$\pm$3   &455$\pm$26   \\
		Walker, walk       &682$\pm$54   & 774$\pm$136                        &538$\pm$194   &760$\pm$90   \\
		Ball in cup, catch &857$\pm$8   &928$\pm$35                         &943$\pm$27   &902$\pm$63   \\ \midrule
		Mean               &749   & 830                        &743   &770   \\
		Median             &786   & 870                        &811   &845   \\ \bottomrule& & 
	\end{tabular}
	\label{table:ablation on predictor depth}
\end{table*}

Table \ref{table:ablation on sequence length} shows the effect of sequence length on model performance in the ablation experiments of continuous control. We use a sequence length of 8 as the baseline. When the sequence length is set to 16, the corresponding block size is multiplied by 2 to ensure a fair comparison as much as possible.

\begin{table*}[h]
	\centering
	\caption{Ablation experiments of sequence length. Results are based on 3 random seeds.}
	\begin{tabular}{lcccc}
		\toprule
		Environment        & 4           &      8 (MPR)      &     16      &     24     \\ \midrule
		Finger, spin       & 635$\pm$26  & 942$\pm$60  & 766$\pm$186 & 584$\pm$11 \\
		Cartpole, swingup  & 850$\pm$13  & 863$\pm$15  & 816$\pm$26  & 808$\pm$28 \\
		Reacher, easy      & 925$\pm$65  & 935$\pm$17  & 919$\pm$56  & 823$\pm$64 \\
		Cheetah, run       & 472$\pm$15  & 538$\pm$43  & 502$\pm$38  & 536$\pm$13 \\
		Walker, walk       & 606$\pm$244 & 774$\pm$136 & 718$\pm$131 & 441$\pm$12 \\
		Ball in cup, catch & 934$\pm$8   & 928$\pm$35  & 928$\pm$53  & 934$\pm$24 \\ \midrule
		Mean               & 737         &     830     &     775     &    688     \\
		Median             & 845         &     870     &     836     &    685\\
		\bottomrule& 
	\end{tabular}
	\label{table:ablation on sequence length}
\end{table*}

The performance of some alternative designs in MPR on the DMControl benchmark is investigated.
All experimental results are shown in Table \ref{table:alternative designs}.
The performance of alternative designs in MPR on the DMControl benchmark is investigated. All experimental results are shown in Table \ref{table:alternative designs}. After adding action sequence information, the average score of MPR decreased by 7.9\% (830$\rightarrow$769). When setting $R$ to 2, the average score of MPR-r (786) decreased by 5.6\% compared to MPR (830), but still outperformed MLR (772.8) \cite{yu2022mask}. This suggests that in continuous control experiments, action sequences do not assist MPR in understanding tasks. Providing multi-view information of images may increase the sequence length, leading to a decrease in performance. Therefore, the performance of MPR-ra is not expected to be better. Replacing the loss function with mean squared loss, MPR-s showed a performance decrease of 11.9\% compared to MPR, indicating that in MPR or reinforcement learning agents with similar auxiliary tasks, it is unnecessary to precisely reconstruct embeddings in the latent space. Setting $R$ to 2 in MPR-s to form MPR-sr resulted in a 3.2\% improvement in model performance compared to MPR-s. However, adding action sequences to both MPR-s and MPR-sr led to a decrease in model performance. Directly masking features, the performance of MPR-fea was not as good as masking images.

\begin{table*}[t]
	\centering
	\caption{
		Ablation experiments on action token, $R$ , mean squared loss, and masking features. MPR-a denotes adding action tokens to the model, MPR-r denotes $R=2$. MPR-ra denotes adding action tokens based on MPR-r. MPR-s represents changing the model's loss function to mean squared loss. MPR-sa represents adding action tokens based on MPR-s. MPR-sr denotes setting $R=2$ based on MPR-s, MPR-sra denotes adding action token sequences based on MPR-sr, MPR-fea represents masking features.}
	\resizebox{\textwidth}{!}{
		\begin{tabular}{lccccccccc}
			\toprule
			Environment        & MPR-a & MPR-r & MPR-ra & MPR-s & MPR-sa & MPR-sr & MPR-sra & MPR-fea&      MPR     \\ \midrule
			Finger, spin       &        896$\pm$54         &903$\pm$106                       &         896$\pm$115          &792$\pm$260                        &          859$\pm$34           &833$\pm$182                           &           872$\pm$169            &614$\pm$30 & 942$\pm$60  \\
			Cartpole, swingup  &        796$\pm$162        &826$\pm$20                      &          828$\pm$66          &816$\pm$85                         &          841$\pm$25           &837$\pm$32                           &            840$\pm$41            &790$\pm$57 & 863$\pm$15  \\
			Reacher, easy      &        848$\pm$74         &919$\pm$66                       &         903$\pm$105          &911$\pm$73                       &          841$\pm$146          &866$\pm$167                           &            834$\pm$86            &833$\pm$113 & 935$\pm$17  \\
			Cheetah, run       &        488$\pm$46         &468$\pm$63                       &          467$\pm$37          &501$\pm$7                         &          415$\pm$31           &512$\pm$43                           &            513$\pm$50            &471$\pm$12 & 538$\pm$43  \\
			Walker, walk       &        697$\pm$277        &702$\pm$153                       &         618$\pm$193          &485$\pm$56                        &          371$\pm$27           & 724$\pm$172                          &            573$\pm$80            &601$\pm$43 & 774$\pm$136 \\
			Ball in cup, catch &        889$\pm$60         &899$\pm$31                       &         890$\pm$104          &945$\pm$21                        &          956$\pm$106          &823$\pm$138                           &           862$\pm$128            &875$\pm$95 & 928$\pm$35  \\ \midrule
			Mean               &            769            &786                       &             767              &742                        &714                               &766                           &               749                &697 &     830     \\
			Median             &            848            &835                       &             816              &807                        & 856                              &827                           &               775                & 635&     870     \\ \bottomrule& & 
		\end{tabular}
	}
	\label{table:alternative designs}
\end{table*}


\clearpage
\section{Hyper-parameters}
We present all hyperparameters used for continuous control and discrete control separately in Table \ref{table:hyprtparameters for dmc} and Table \ref{table:hyprtparameters for atari}, respectively. In continuous control, we set the block size to 8 for \textit{Finger, spin; Walker, walk; and Ball in cup, catch}, and for \textit{Cheetah, run; Cartpole, swingup; Reacher, easy}, we set the block size to 1, 4, and 3, respectively. In continuous control, we differ from the original implementation of MLR in only a few parameters. First, we set the learning rate to 128. Second, we set the learning rate to $1e-4$ for all tasks. 
Third, for all atari games, we set the weight of the MPR loss to 1.

\begin{table*}[h]
	\centering
	\caption{An overview of hyper-parameters in DMControl experiments.}
	\label{table:hyprtparameters for dmc}
	\begin{tabular}{ll}
		\toprule
		\textbf{Hyperparameter} & \textbf{Value}\\
		\midrule
		Obervation image size & (100, 100)\\
		Down sampling image Size & (84, 84)\\
		Frame stack& 3\\
		Augmentation& Random crop and random intensity\\
		Replay buffer size & 100,000\\
		Initial exploration steps & 1,000\\
		Action repeat & 2 \textit{Finger, spin} and \textit{Walker, walk}; 8 \textit{Cartpole, swingup}; 4 \textit{oherwise}\\
		Evaluation episodes & 10\\
		Optimizer & Adam\\
		$(\beta_1, \beta_2)\rightarrow (\theta_{f}, \theta_{g}, \theta_{h})$& (0.9, 0.999)  \\
		$ \beta_1, \beta_2)\rightarrow (a)$ (temperature in SAC)&(0.5, 0.999)  \\
		Learning rate ($\theta_{f}$, $\theta_{\phi}$) & $2e-4$ \textit{Cheetah, run}; $1e-3$ \textit{otherwise}  \\ 			
		Learning rate ($\theta_{f}$, $\theta_{g}$, $\theta_{h}$)& $1e-4$ \textit{Cheetah, run}; $5e-4$ \textit{otherwise}    \\ 	
		Learning rate warmup ($\theta_{f}, \theta_{g}$, $\theta_{h}$)& 6000 steps  \\ 	
		Learning rate ($\alpha$)& $1e-4$   \\ 	
		Batch size for policy learning & 512\\ 
		Batchsize for auxiliary task &128\\
		Q-function EMA $\tau$ & 0.99\\
		Critic target update freq & 2\\
		Discount factor & 0.99\\
		Initial temperature & 0.1\\
		Target network update period & 1\\
		Target network EMA $tau$& 0.9 \textit{Walk, walk } 0.95 \textit{otherwise}\\
		State representation dimension $d$ & 50\\
		\midrule
		\textbf{MPR Specific Hyperparameters}&\\
		\midrule
		Weight of MPR loss $\lambda$ & 1\\
		Mask size & (40, 40)\\
		Sequength length $T$ & 8\\
		Decoder depth $L$&2\\
		$R$ & 1\\
		Block Size & \parbox{10cm}{8 \textit{Finger, spin; Walker, walk; Ball in cup, catch}; 1 \textit{Cheetah, run}; 4 \textit{Cartpole, swingup}; 2 \textit{Reacher, easy}}\\
		\bottomrule
	\end{tabular}
\end{table*}

\begin{table*}[h]
	\centering
	\caption{An overview of hyper-parameters in Atari experiments.}
	\label{table:hyprtparameters for atari}
	\begin{tabular}{ll}
		\toprule
		\textbf{Hyperparameter} &\textbf{Value}\\
		\midrule
		Observation downsampling & (84, 84)\\
		Augmentation &Random crop and random intensity\\
		Frame stack &4\\
		Gray-scaling &True\\
		Action repeat &4\\
		Training steps &100,000\\
		Max frames per episode &108,000\\
		Replay buffer size &100,000\\
		Minimum replay size for sampling &2,000\\
		Mini-batch size &128\\
		Optimizer &Adam\\
		Optimizer learning rate &$1e-4$\\
		$(\beta_1, \beta_2)$& (0.9, 0.999)\\
		Optimizer $\eta$&0.00015\\
		Max gradient norm &10\\
		Update &Distributional Q\\
		Dueling &True\\
		Support of Q-distribution &51 bins\\
		Discount factor &0.99\\
		Reward clipping Frame stack &[-1,1]\\
		Priority exponent&0.5\\
		Priority correction&0.4$\rightarrow$1\\
		Exploration&Noisy nets\\
		Noisy nets parameter&0.5\\
		Evaluation trajectories &100\\		
		Replay period every &1 step\\
		Updates per step&2\\
		Multi-step return length &10\\
		Q network:channels &32, 64, 64\\
		Q network: filter size &8$\times$8, 4$\times$4, 3$\times$3\\
		Q network: stride&4, 2, 1\\
		Q network:hidden units&256\\
		Target network update period&1\\
		$\tau$(EMA coefficient)&0\\
		\midrule
		\textbf{MPR Specific Hyerparameters}&\\
		\midrule
		Weight of MPR loss $\lambda$&1\\
		Mask size &(32$\times$32)\\
		Sequength length $T$ & 16\\
		Decoder depth $L$&2\\
		$R$ & 2\\
		Block size & 8\\
		\bottomrule
	\end{tabular}
\end{table*}

\end{document}